\documentclass[letterpaper]{article} 
\usepackage[draft]{aaai2026}  
\usepackage{times}  
\usepackage{helvet}  
\usepackage{courier}  
\usepackage[hyphens]{url}  
\usepackage{graphicx} 
\usepackage{natbib}  
\usepackage{caption} 
\usepackage{algorithm}
\usepackage{algorithmic}
\usepackage{amsmath}
\usepackage{booktabs}

\title{Images Speak Louder Than Scores: \\ Failure Mode Escape for Enhancing Generative Quality}

\author{%
  Jie Shao \quad Ke Zhu \quad Minghao Fu \quad Guo-hua Wang \quad Jianxin Wu\thanks{Corresponding author.}
}
\affiliations{
  National Key Laboratory for Novel Software Technology, Nanjing University, China\\
  School of Artificial Intelligence, Nanjing University, China\\
  \texttt{\{shaoj, zhuk, fumh, wanggh\}@lamda.nju.edu.cn, wujx2001@nju.edu.cn} \\
}

\begin{document}

\maketitle

\begin{abstract}
Diffusion models have achieved remarkable progress in class-to-image generation. However, we observe that despite impressive FID scores, state-of-the-art models often generate distorted or low-quality images, especially in certain classes. This gap arises because FID evaluates global distribution alignment, while ignoring the perceptual quality of individual samples. We further examine the role of CFG, a common technique used to enhance generation quality. While effective in improving metrics and suppressing outliers, CFG can introduce distribution shift and visual artifacts due to its misalignment with both training objectives and user expectations. In this work, we propose FaME, a training-free and inference-efficient method for improving perceptual quality. FaME uses an image quality assessment model to identify low-quality generations and stores their sampling trajectories. These failure modes are then used as negative guidance to steer future sampling away from poor-quality regions. Experiments on ImageNet demonstrate that FaME brings consistent improvements in visual quality without compromising FID. FaME also shows the potential to be extended to improve text-to-image generation.

\end{abstract}

\section{Introduction}

Diffusion models have recently achieved remarkable success in image generation~\cite{sdxl,sd3,labs2025flux}, video generation~\cite{sora,wan,seedance}, and even language generation~\cite{llada,mmada}. Their core idea is to synthesize the target output through a step-by-step denoising and refinement process starting from Gaussian noise~\cite{ddpm,song2020score}. Currently, models that predict noise~\cite{song2021denoising}, score functions~\cite{score0}, or velocity fields~\cite{lipman2022flow} have been extensively studied, and Transformer-based architectures~\cite{dit,sit} have been widely adopted. Another critical component is the solver~\cite{song2021denoising,dpm-solver,dpm-solver++}, which numerically solves the underlying SDE/ODE formulations. With continuous advancements in both model architecture and solver design, diffusion models are now capable of generating high-quality images and videos both rapidly and effectively.

Among various benchmarks, the class-to-image (C2I) generation task on ImageNet has become an important experimental setting that offers valuable guidance~\cite{vqgan,rombach2022high}. This task has also witnessed significant progress in recent years. 

This paper emphasizes two elements that are often considered fundamental (sometimes even taken for granted) in the generation of C2I: \emph{distribution-based evaluation metrics}~\cite{fid,is} and \emph{classifier-free guidance} (CFG)~\cite{cfg}.

Distribution-based metrics, such as FID, have become the most widely used indicators for evaluating the performance of C2I models. Recently, FID scores on ImageNet have shown remarkable improvements, in some cases even surpassing those computed on the real images from the validation set~\cite{var,repa,vavae}. However, when we examine the actual outputs of state-of-the-art generative models, we often observe unrealistic, distorted, or visually unappealing images. This inconsistency arises mainly because metrics like FID embed images into a feature space and compute the distance between overall distributions, but \emph{does not assess the quality of individual samples}. In contrast, when we evaluate generated images using image quality assessment (IQA) models~\cite{qalign,hyperiqa}, a different picture emerges: (a) There is a clear imbalance in generative quality across classes: While some classes yield high-quality images, many others frequently produce low-quality or distorted results. (b) A significant perceptual gap remains between generated and real images—one that is more aligned with human judgment than what FID or IS scores would suggest.

We argue that this deficiency in generation quality stems partly from the model’s capacity—namely, its limited ability to consistently produce high-quality images—and \emph{partly from the use of CFG}, which is widely adopted in image generation. In C2I generation, CFG typically guides the sampling process by incorporating unconditional generation, thereby helping to avoid outliers and improve visual quality. CFG plays a crucial role in boosting both distribution-based metrics and perceptual quality. In fact, some models cannot generate satisfactory outputs without CFG. However, CFG introduces inherent issues. It inevitably leads to distributional shift and can overshoot the unintended distribution, resulting in unexpected artifacts such as distortion or blurriness~\cite{autoguidance,guidance-interval}. This problem arises because the CFG mechanism is not fully aligned with either the training objective—which aims to approximate the true data distribution in both conditional and unconditional settings—or the ultimate goal from a user perspective: generating visually high quality images.

Our goal is to improve image quality without retraining the generative model or increasing inference time (unlike test-time scaling methods~\cite{tts}). To this end, we propose an enhancement based on CFG. We leverage an IQA model to identify the lowest-quality generated samples and store their sampling trajectories in a small memory pool. These trajectories represent \emph{failure modes to escape from}. During sampling, we reuse the score functions from these failure cases as negative guidance, effectively steering the generation process away from low-quality regions and toward higher-quality outputs. The entire process is \emph{training-free}, incurs \emph{negligible memory overhead}, and does \emph{not increase inference time}, as it only requires storing a small set of trajectories without recomputation. We refer to this method as \textbf{Failure Mode Escape (FaME)}.

Our contributions are summarized as follows:
\begin{itemize}
    \item We identify a critical issue in current image generation research: excessive focus on distribution-based metrics (e.g., FID) while neglecting perceptual image quality. We highlight the class-wise quality imbalance and the persistent gap between generated and real images.
    
    \item We propose FaME, a training-free and inference-efficient method that leverages failure-mode score functions as negative guidance. FaME significantly improves image quality to state-of-the-art levels, with clear visual enhancements compared to standard CFG.
\end{itemize}

\section{Related Work}

\textbf{Diffusion Models.} Diffusion models~\cite{yang2023diffusion} have recently emerged as state-of-the-art generative models, achieving impressive results in image generation tasks. These models generate samples by reversing a noise-adding forward process, typically modeled as either a stochastic differential equation (SDE)~\cite{song2020score} or an ordinary differential equation (ODE)~\cite{song2021denoising}.  Unlike traditional GAN-based methods~\cite{goodfellow2014generative, arjovsky2017wasserstein, brock2018large}, diffusion models rely on score-matching techniques~\cite{hyvarinen2005estimation} to approximate the gradient of the log-density of the data distribution. Improved variants such as EDM~\cite{karras2022elucidating} and LDM~\cite{rombach2022high} have further enhanced sampling efficiency and generation quality. Recently, flow matching models~\cite{lipman2022flow,dao2023flow} have also gained attention. These models directly match data distributions through continuous normalizing flows, leveraging optimal transport theories and demonstrating promising performance in terms of faster sampling and stable training dynamics~\cite{sd3,labs2025flux}.

\textbf{Guidance Techniques.} Guidance methods have been extensively studied to enhance controllability in conditional generation. Initially, classifier guidance (CG)~\cite{cg} leveraged gradients from separately-trained classifiers to direct sampling towards target classes but introduced additional complexity. Classifier-free guidance (CFG)~\cite{cfg} emerged as a simpler alternative by combining conditional and unconditional score estimates, significantly improving sample fidelity and semantic alignment. Recent studies identified limitations of CFG, noting that high guidance scales may distort distributions and degrade quality~\cite{guidance-interval, autoguidance}. Methods such as adaptive masking of low-confidence regions~\cite{li2025adaptive}, spatially-sensitive guidance weighting~\cite{shen2025spatial}, guidance interval limitation~\cite{guidance-interval}, and over-saturation artifact correction~\cite{sadat2024eliminating} have been proposed to mitigate these issues.

\textbf{Generative Quality Metrics.} Accurately evaluating generative models remains challenging. Traditional metrics like Fréchet Inception Distance (FID)~\cite{fid} and Inception Score (IS)~\cite{is} are often combined to jointly measure sample fidelity and diversity. Newer distributional measures such as Kernel Inception Distance (KID)~\cite{kid} and Precision \& Recall for Distributions~\cite{prec_recall} further disentangle quality from coverage. Pixel-level fidelity measures like PSNR~\cite{psnr} quantify raw signal similarity, whereas perceptual similarity metrics such as SSIM~\cite{ssim} and LPIPS~\cite{lpips} better capture structural and semantic difference.

For text-to-image, CLIPScore~\cite{clipscore} evaluates semantic coherence with prompts. More comprehensive suites such as Geneval~\cite{geneval} and DPGBench~\cite{dpgbench} aggregate multiple evaluation tasks for robustness, and human-comparison benchmarks like HRS-Bench~\cite{hrs_bench} and T2I-CompBench~\cite{t2i_compbench} extend assessments across object-level, compositionality, and fairness. Perceptually-aligned methods such as Q-Align~\cite{qalign} leverage multimodal large-language model embeddings to better reflect human judgment.

\section{Background}

\textbf{Diffusion.} Diffusion models generate samples from a data distribution $p(\mathbf{x})$ by progressively reversing a noise-adding process~\cite{ddpm}. The forward process is modeled as a sequence of noised distributions: $p(\mathbf{x}_\sigma) = p(\mathbf{x}) * \mathcal{N}(0, \sigma^2 I)$. The generation process is then formulated as solving an ODE that evolves backward in the noise scale $\sigma$, from a high noise level $\sigma_{\max}$ down to $\sigma = 0$:
\begin{equation}
\frac{d\mathbf{x}_\sigma}{d\sigma} = -\sigma\nabla_{\mathbf{x}_\sigma} \log p(\mathbf{x}_\sigma; \sigma)\,,
\end{equation}
which is solved numerically by estimating the score function $\nabla_{\mathbf{x}} \log p(\mathbf{x}_\sigma; \sigma)$ using a neural network $D_\theta$, trained to denoise perturbed samples~\cite{song2021denoising}.

\begin{table}[t]
	\centering
	\begin{tabular}{l|ccc|c}
		\toprule
		Model & FID$\downarrow$ & Precision$\uparrow$ & Recall$\uparrow$ & $S_{\text{Q}}\uparrow$ \\
		\midrule
		\textcolor[gray]{0.7}{(Real Images)} & \textcolor[gray]{0.7}{1.78} & \textcolor[gray]{0.7}{0.75} & \textcolor[gray]{0.7}{0.67} & \textcolor[gray]{0.7}{2.60} \\
		\midrule
		DiT-XL/2 & 2.27 & 0.83 & 0.57 & 2.30 \\
		SiT-XL/2 & 2.15 & 0.81 & 0.60 & 2.32 \\
		VAR-d30 & 1.97 & 0.81 & 0.61 & 2.35 \\
		\bottomrule
	\end{tabular}
	\caption{Performance of various generative models (with CFG). The adopted metrics include widely used ones such as FID, as well as the IQA metric ($S_{\text{Q}}$). $\uparrow$ ($\downarrow$) means larger (smaller) is better.}
	\label{table: fid}
\end{table}

\begin{figure*}[t]
    \centering
    \includegraphics[width=0.9\linewidth]{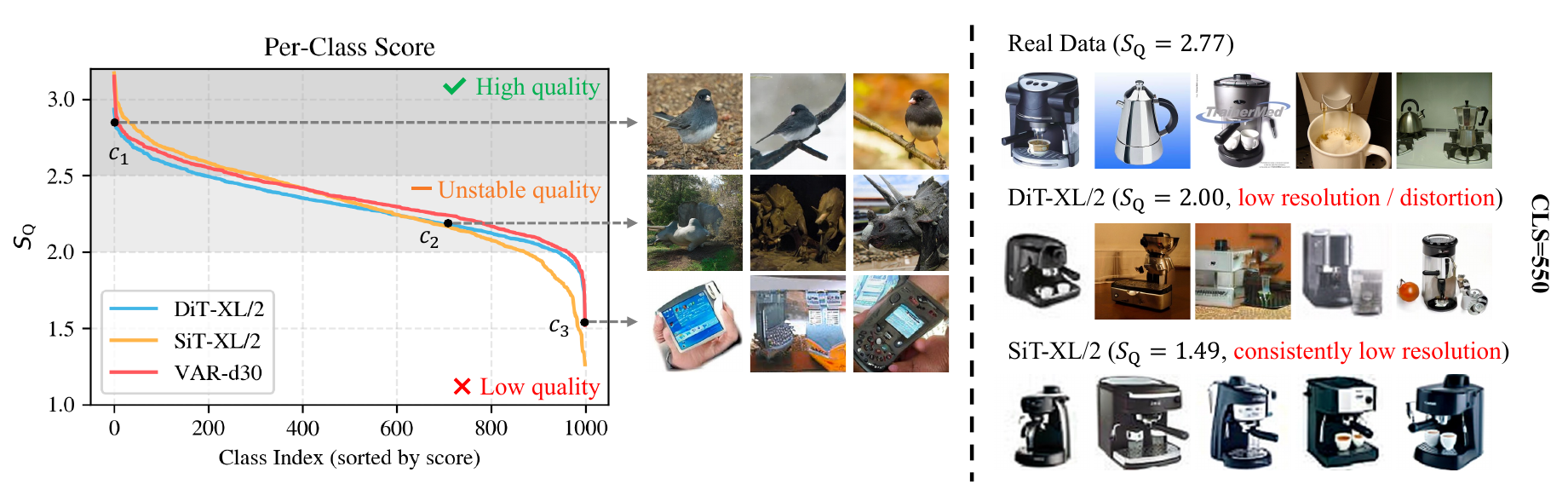}
    \caption{Distribution of $S_{\text{Q}}$ across classes for different generative models, along with representative generation examples. A clear imbalance in per-class generative quality is observed, which can be roughly categorized into three quality tiers (left). Classes with low $S_{\text{Q}}$ scores often exhibit noticeable blurriness or distortion—artifacts that are typically overlooked by distribution metrics such as FID (right). More visualization results are provided in the appendix.}
    \label{fig: qalign}
\end{figure*}

\textbf{Classifier-Free Guidance.} In conditional generation, our aim is to sample from the distribution $p(\mathbf{x} | c; \sigma)$, where $c$ denotes a class label or text prompt. To this end, the denoiser is extended to accept $c$ as an additional input as $D_\theta(\mathbf{x}, \sigma, c)$, and is trained to reconstruct clean samples conditioned on $c$. To further enhance sample fidelity and semantic alignment, researchers adopt classifier-free guidance (CFG)~\cite{cfg}. Following the formulation in~\cite{autoguidance}, the guided denoiser is defined as:
\begin{equation}
D_w(\mathbf{x}, \sigma, \mathbf{c}) = w D_1(\mathbf{x}, \sigma, \mathbf{c}) + (1 - w) D_0(\mathbf{x}, \sigma)\,,
\end{equation}
where $D_0(\mathbf{x}, \sigma)$ is the unconditional denoiser (acting as negative guidance) and $D_1(\mathbf{x}, \sigma, c)$ is the conditional denoiser (providing positive guidance). CFG has proven to be crucial for improving generation quality. With appropriate guidance scaling, diffusion models equipped with CFG have achieved \textit{seemingly impressive} performance on standard benchmarks, even being close to or surpassing real images in FID scores, as illustrated in Table~\ref{table: fid}.

\section{The Image Quality Pitfall}

\textbf{Illusion of FID.} The FID metric in Table~\ref{table: fid} seems to suggest that the generative models are capable of producing images that are as realistic as real images. Is this truly the case? For some classes, yes. However, when we examine these models more closely and attempt to generate images from other classes, this conclusion no longer holds consistently.

To better assess the actual generative quality of these models, we employ image quality assessment (IQA) techniques. Specifically, we sample 50 images per class (50,000 images in total) and evaluate them using Q-Align~\cite{qalign}---a state-of-the-art IQA model~\cite{pyiqa} based on a multimodal large language model. The average scores across all classes are reported in Table~\ref{table: fid} (denoted as $S_{\text{Q}}$). In addition, we visualize the per class distribution of $S_{\text{Q}}$ in Figure~\ref{fig: qalign} to provide deeper insights.

How reliable is $S_{\text{Q}}$, and what does it actually reflect? The distribution in Figure~\ref{fig: qalign} provides a preliminary view of this metric. We roughly categorize the classes into three groups based on the scores and corresponding images:

\begin{itemize}
\item $S_{\text{Q}} > 2.5$: The generated images in these classes appear natural and high-quality. These classes are typically selected for qualitative display in the literature.
\item $S_{\text{Q}} < 2.0$: The generated images often exhibit distortion or blurriness. The overall quality is poor.
\item $2.0 \leq S_{\text{Q}} \leq 2.5$: The quality is inconsistent; some samples are good, others are not.
\end{itemize}

For example, class 550 yields low $S_{\text{Q}}$ score for DiT-XL/2 and SiT-XL/2, as shown in Figure~\ref{fig: qalign}. The score $S_{\text{Q}}$ is directly indicative of perceptual image quality, and even a 0.1 increase in $S_{\text{Q}}$ typically corresponds to a noticeable improvement in visual fidelity.

Despite achieving competitive FID scores, these models still produce low-quality and unappealing images for many classes. A substantial number of classes with $S_{\text{Q}} < 2.5$ consistently yield unsatisfactory results, exposing a significant gap in class-wise generative quality that is often obscured by aggregate metrics such as FID. 

This highlights two key limitations of current generative models: (a) a pronounced imbalance in image quality between classes, masked by general metrics; (b) while FID has improved rapidly, progress in image quality (as reflected in $S_{\text{Q}}$) has seen much slower and has attracted less attention.

\begin{figure*}[t]
    \centering
    \includegraphics[width=0.9\linewidth]{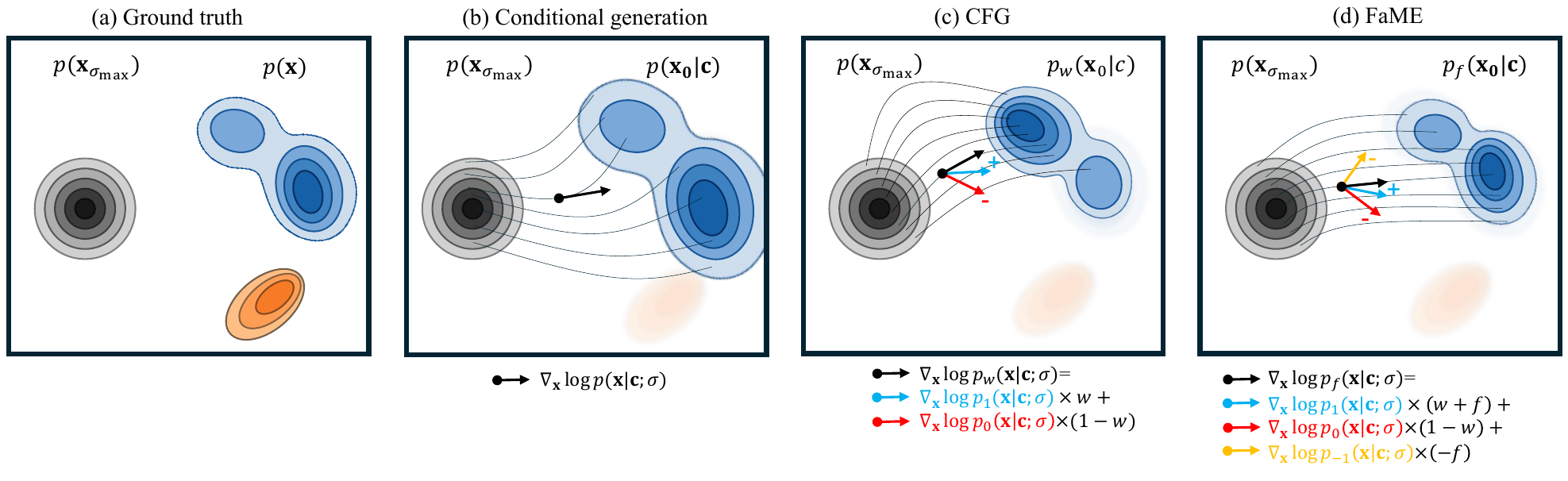}
    \caption{Visualization and Comparison of Sampling Methods. (a) The initial Gaussian distribution and the target ground-truth distribution. (b) Conditional generation tends to produce out-of-distribution samples, resulting in outliers. (c) CFG introduces the unconditional score function as negative guidance, reducing outliers but causing distribution shift and over-emphasis. (d) our FaME provides alternative negative guidance that suppresses over-sampling of sparse, low-quality samples, correcting the distribution and improving generation quality.  In the figure, plus and minus signs indicate the positive and negative score directions. Typical values for $w$ and $f$ are 1.5 and 0.1.
    }
    \label{fig: fame}
\end{figure*}

\textbf{Dilemma of CFG.} We aim to design a training-free method to directly refine the image generation quality. Since classifier-free guidance (CFG) has been shown to be highly effective in improving metrics such as FID and IS, we turn to guidance-based techniques to address this problem.

Recall that CFG was originally proposed to solve a key challenge in conditional generation: the conditional network $D_\theta$ often fails to fully enforce the conditioning information, leading to unappealing outliers~\cite{autoguidance}. As illustrated in Figure~\ref{fig: fame}(b), such outliers are difficult to avoid for each class due to the weak controllability of the conditioning signal. CFG suppresses outliers at the cost of diversity, aiming to enforce stronger conditional consistency. Compared to the original conditional distribution $p(\mathbf{x} | \mathbf{c}; \sigma)$ and its associated score function $\nabla_\mathbf{x} \log p(\mathbf{x} | \mathbf{c}; \sigma)$, the distribution and score used in CFG are as follows:
\begin{align}
 p_w(\mathbf{x} | \mathbf{c}; \sigma) 
& \propto 
p_0(\mathbf{x} | \mathbf{c}; \sigma)
\left[
\frac{p_1(\mathbf{x} | \mathbf{c}; \sigma)}{p_0(\mathbf{x} | \mathbf{c}; \sigma)}
\right]^w, \\
\nabla_{\mathbf{x}} \log p_w(\mathbf{x} | \mathbf{c}; \sigma)
&= w\nabla_{\mathbf{x}} \log p_1(\mathbf{x} | \mathbf{c}; \sigma) \notag \\
& \quad\,\, + (1 - w) \nabla_{\mathbf{x}} \log 
p_0(\mathbf{x} | \mathbf{c}; \sigma),
\end{align}
where $p_1(\mathbf{x} | \mathbf{c}; \sigma)$ denotes the conditional distribution, typically estimated by $D_1$, while $p_0(\mathbf{x} | \mathbf{c}; \sigma)$ represents the unconditional distribution estimated by $D_0$. We can interpret CFG more intuitively by viewing $\nabla_{\mathbf{x}} \log p_1(\mathbf{x} | \mathbf{c}; \sigma)$ as a positive update direction and $\nabla_{\mathbf{x}} \log p_0(\mathbf{x} | \mathbf{c}; \sigma)$ as a negative one. This guidance mechanism enhances the conditional signal while suppressing outliers.

Despite its wide adoption and strong improvements in metrics such as FID, CFG has been criticized for inducing distribution shifts and leading to distorted sampling trajectories~\cite{guidance-interval, autoguidance}. As illustrated in Figure~\ref{fig: fame}(c), CFG can alter the sampling distribution, making dense regions sparse and vice versa. This misalignment can cause the model to over-sample low-quality areas, ultimately degrading image quality.

\begin{table}[t]
	\centering
	\begin{tabular}{cc|cc|cc}
		\toprule
		$w$ & $S_{\text{Q}}\uparrow$ & $w$ & $S_{\text{Q}}\uparrow$ & $w$ & $S_{\text{Q}}\uparrow$  \\
		\midrule
		1.0  & 2.03 & 1.3 & 2.12 & 2 & 1.99 \\
		1.1  & 2.09 & 1.5 & 2.01 & 3 & 1.92 \\
		1.2  & 2.11 & 1.8 & 2.00 & 5 & 1.54\\
		\midrule
		Real & 2.77 & $\text{FaME}_\mathbf{c}$ & 2.46 & FaME &  2.43     \\
		\bottomrule
	\end{tabular}
	\caption{Case study comparison of CFG and FaME (our method) in enhancing generation quality. $w$ denotes the CFG scale. Compared to CFG, our method (both the original version $\text{FaME}_\mathbf{c}$ and the refined and accelerated version FaME) achieves significantly better results even with the default hyperparameter.}
\label{table: cfg}
\end{table}

This raises a natural question: can we achieve balance and improve image quality by more finely adjusting the CFG scale? To explore this, we take class 550 (\textsc{espresso maker}) as a case study. We vary the CFG scale $w$ and sample 50 images using DiT-XL/2, evaluating the corresponding generation quality using Q-Align. The results, shown in Table~\ref{table: cfg}, suggest that even with careful tuning, the quality improvements remain limited. The model either suffers from outliers due to weak conditional control or from distribution shifts introduced by CFG. In either case, the  generation quality remains difficult to improve.

\section{FaME: Failure Mode Escape}

We closely examined the samples generated with and without CFG. Although CFG significantly reduces the number of meaningless, confusing, or class-irrelevant images, an unexpected phenomenon arises. For example, in class 550 (\textsc{espresso maker}), although low-quality and blurry images—resembling those shown in the third row of Figure~\ref{fig: qalign} (right)—do exist in the training set, they account for only a small fraction. But, after applying CFG, the generative model tends to overproduce images of this degraded style. This suggests that CFG may inadvertently over-sample from low-density regions of the data distribution, which in turn restricts further improvements in image quality. We aim to mitigate the distribution shift introduced by CFG and thereby enhance generation quality.

\textbf{Score Function for Failure Samples.} Our goal is to steer both the sampling trajectory and the final distribution away from failure samples and low-frequency modes. Let $p_{-1}(\mathbf{x} | \mathbf{c}; \sigma)$ denotes the distribution of such undesired samples. In contrast to the CFG-induced distribution $p_{w}(\mathbf{x} | \mathbf{c}; \sigma)$, our proposed distribution is designed to be
\begin{equation}
p_f(\mathbf{x} | \mathbf{c}; \sigma) \propto 
\frac{\left[p_1(\mathbf{x} | \mathbf{c}; \sigma)\right]^w}{\left[p_0(\mathbf{x} | \mathbf{c}; \sigma)\right]^{(w-1)}}
 \left[
\frac{p_1(\mathbf{x} | \mathbf{c}; \sigma)}{p_{-1}(\mathbf{x} | \mathbf{c}; \sigma)}
\right]^f \,,
\end{equation}
which suppresses both outliers and failure modes. The main challenge is that the distribution $p_{-1}$ is unknown and even harder to estimate than the standard conditional distribution. Unlike $p_1$ and $p_0$, which can be approximated relatively easily using score-based models $D_1$ and $D_0$, accurately estimating $p_{-1}$ would require training a separate model $D_{-1}$, which is not only resource intensive but also deviates from our goal. Therefore, we consider an alternative approach.

\begin{figure}[t]
    \centering
    \includegraphics[width=0.9\linewidth]{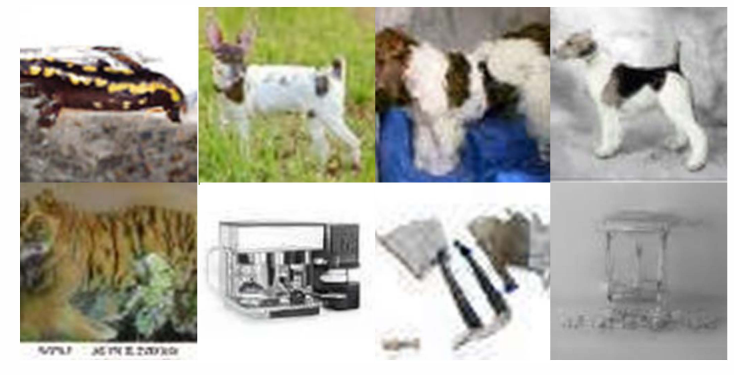}
    \caption{Example failure mode generation results. FaME leverages the sampling trajectories of these low-quality images as negative guidance to improve generation.}
    \label{fig: fail}
\end{figure}

For a given class $\mathbf{c}$, we can easily sample $n_{\mathbf{c}}$ data points, evaluate them using the quality score $S_\text{Q}$, and rank the results. By selecting the lowest-scoring $n_{f,\mathbf{c}}$ samples, we obtain a set that we believe approximates the samples drawn from the failure distribution $p_{-1}$. Examples of samples treated as failure modes are shown in Figure~\ref{fig: fail}. Instead of explicitly estimating this distribution, we propose using the estimated score function of these failure samples to guide the model away from them.  Since we perform generation through ODE-based iterations, the sampling trajectory is fully determined once the initial noise is fixed. Thus, we record the initial noise vectors corresponding to the failure samples and approximate the score function for the failure samples as:
\begin{align}
\nabla_{\mathbf{x}} \log p_{-1}(\mathbf{x} | \mathbf{c}; \sigma) &\approx \nabla_{\mathbf{x}} \log p(\mathbf{x} | \mathbf{c}; \sigma_{f,\mathbf{c}}) \,, \\
\sigma_{f,\mathbf{c}} &\sim \text{Uniform}(\Sigma_{f,\mathbf{c}}), \ |\Sigma_{f,\mathbf{c}}| = n_{f,\mathbf{c}} \,.
\end{align}

Our final score function is then defined as
\begin{align}
&\nabla_{\mathbf{x}} \log p_f(\mathbf{x} | \mathbf{c}; \sigma)
= (w + f) \nabla_{\mathbf{x}} \log p_1(\mathbf{x} | \mathbf{c}; \sigma) \\
&\quad + (1 - w) \nabla_{\mathbf{x}} \log p_0(\mathbf{x} | \mathbf{c}; \sigma) 
- f \nabla_{\mathbf{x}} \log p_{-1}(\mathbf{x} | \mathbf{c}; \sigma). \notag
\end{align}
Given the approximation $\nabla_{\mathbf{x}} \log p(\mathbf{x}; \sigma) \approx \frac{D_\theta(\mathbf{x}; \sigma) - \mathbf{x}}{\sigma^2}$,
our proposed \textbf{FaME} (Failure Mode Escape) method can be formulated in terms of model outputs as
\begin{align}
D_f(\mathbf{x}, \sigma, \mathbf{c}) 
&= (w + f) D_1(\mathbf{x}, \sigma, \mathbf{c}) \notag\\
&+ (1 - w) D_0(\mathbf{x}, \sigma) 
- f D_1(\mathbf{x}, \sigma_{f,c}, \mathbf{c})\,. 
\end{align}

\begin{table}[t]
	\centering
	\begin{tabular}{cc|ccc|c}
		\toprule
		$f$ & $\tau$ & FID$\downarrow$ & Precision$\uparrow$ & Recall$\uparrow$ & $S_\text{Q}$$\uparrow$ \\
		\midrule
		0.0 & 0.0   & 2.21 & 0.82 & 0.57 & 2.30 \\
		\midrule
		0.02 & 0.3  & 2.18 & 0.84 & 0.57 & 2.52 \\
		0.02 & 0.5  & 2.39 & 0.84 & 0.56 & 2.54 \\
		0.05 & 0.5  & 3.16 & 0.84 & 0.54 & 2.61 \\
		\bottomrule
	\end{tabular}
	\caption{Model performance under different $f$ and $\tau$ settings. We choose $f = 0.02$ and $\tau = 0.3$ as the default setting, as it does not degrade FID and significantly improves the quality score $S_\text{Q}$.}
	\label{tab: hyperparameter}
\end{table}

\begin{figure*}[t]
    \centering
    \includegraphics[width=\linewidth]{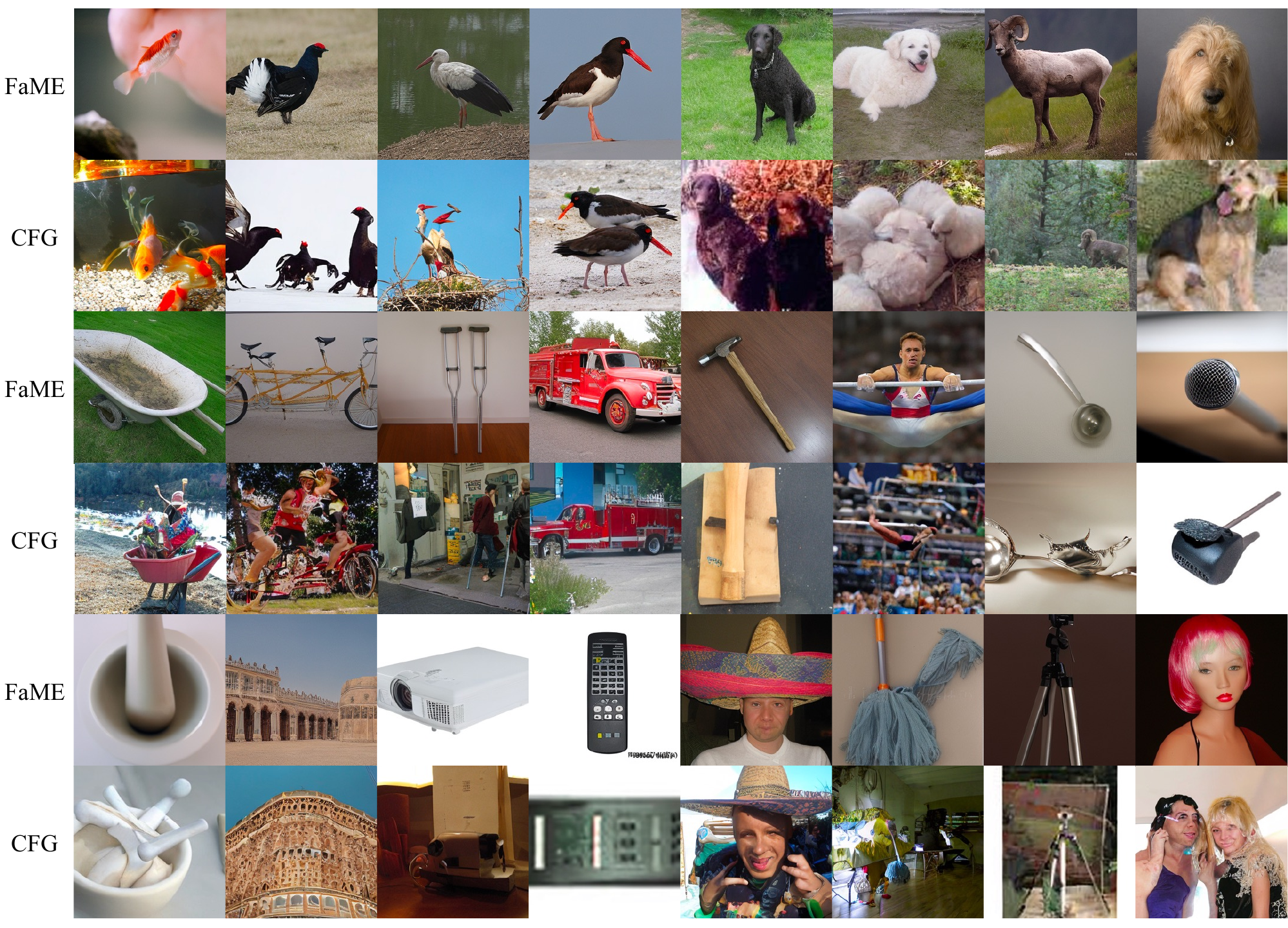}
    \caption{Visual comparison between FaME and baseline results. We adopt DiT-XL/2 with CFG as the baseline. It is clear that the baseline often generates distorted or visually unappealing images, while our method significantly improves generation quality across all classes.The corresponding images generated using FaME and CFG share \emph{the same initial noise}.}
    \label{fig: visual-compare}
\end{figure*}

\begin{table*}[t]
	\centering
	\begin{tabular}{l|cccc|cc}
		\toprule
		Model & FID$\downarrow$ & IS$\uparrow$ & Precision$\uparrow$ & Recall$\uparrow$ & $S_\text{Q}\uparrow$ & $S_\text{H}\uparrow$ \\
		\midrule
		\textcolor[gray]{0.7}{(Real Data)} & \textcolor[gray]{0.7}{1.78} & \textcolor[gray]{0.7}{236.90} & \textcolor[gray]{0.7}{0.75} & \textcolor[gray]{0.7}{0.67} & \textcolor[gray]{0.7}{2.60} & \textcolor[gray]{0.7}{0.63} \\
		\midrule
		DiT-XL/2                 & 2.27 & 278.24 & 0.83 & 0.57 & 2.30 & 0.59 \\
		SiT-XL/2                 & 2.15 & 258.09 & 0.81 & 0.60 & 2.32 & 0.57 \\
		VAR-d30                  & 1.97 & 334.70 & 0.81 & 0.61 & 2.35 & 0.60 \\
		REPA                     & 1.42 & 305.70 & 0.80 & 0.65 & 2.36 & 0.61 \\
		LightningDiT             & 1.35 & 295.30 & 0.79 & 0.65 & 2.26 & 0.58 \\
		\midrule
		DiT-XL/2+FaME            & 2.18 & 281.26 & 0.84 & 0.57 & 2.52 & 0.63 \\
		SiT-XL/2+FaME            & 2.17 & 254.92 & 0.82 & 0.58 & \textbf{2.53} & \textbf{0.63} \\
		\bottomrule
	\end{tabular}
	\caption{Comparison of different models on FID, IS, Precision, Recall, $S_\text{Q}$, and $S_\text{H}$ metrics. Although recent methods achieve impressive performance on traditional metrics such as FID, our FaME enables baseline models to reach SOTA quality scores ($S_\text{Q}$ and $S_\text{H}$) without compromising FID.
	}
	\label{tab: result}
\end{table*}

\begin{figure*}
    \centering
    \includegraphics[width=0.95\linewidth]{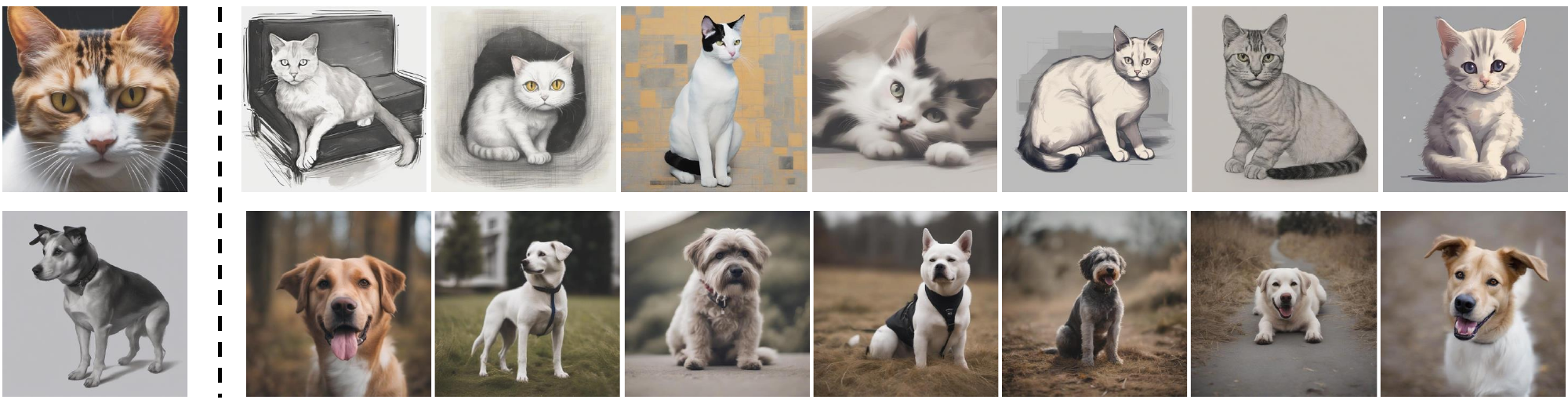}
    \caption{Text-to-image generation results using FaME. The left side shows the negative samples, while the right side presents the corresponding outputs. FaME effectively guides the model to avoid generating images similar to the negative examples.}
    \label{fig: t2i}
\end{figure*}

One potential approach to accelerate the inference speed is to store not only the initial noise, but also the intermediate outputs of the model along the entire trajectory. However, this would significantly increase the storage requirements. To balance efficiency and storage cost, we propose a simple yet effective compromise: instead of storing $n_{f,\mathbf{c}}$ trajectories for each class, we store a total of $n_f$ trajectories globally across all classes (default is 8). During inference, we randomly select one trajectory from this global pool, regardless of class label, and use it for sampling. This reduces the memory footprint while maintaining the inference speed of FaME. Specifically, we implement the following strategy:
\begin{align}
\nabla_{\mathbf{x}} \log p_{-1}(\mathbf{x} | \mathbf{c}; \sigma) &\approx \nabla_{\mathbf{x}} \log p(\mathbf{x} | \mathbf{c'}; \sigma_f), \\
\sigma_f, \mathbf{c}' &\sim \text{Uniform}(\Sigma_f), \ |\Sigma_f| = n_f.
\end{align}
Now FaME achieves an inference speed identical to that of CFG, and the storage of the $n_{f}$ trajectory can be neglected. As shown in Table~\ref{table: cfg}, FaME refers to our refined version, which results in only a negligible drop in $S_\text{Q}$---a trade-off that is entirely acceptable. Therefore, we choose not to store the per-class negative trajectory by default.

FaME also introduces a new hyperparameter $\tau$, inspired by~\citet{guidance-interval}, which determines from which step the FaME term is applied during denoising. $\tau$ controls the diversity of the generated images. By default, we set $\tau = 1.0$, which means that the FaME term is applied from the very beginning of the denoising process. In the context of DDPM, setting $\tau = 1.0$ corresponds to adding the FaME term at step 1000, which also marks the start of the denoising process. More details on how $f$ and $\tau$ are chosen are provided in the experimental section.

\section{Experiments}

\textbf{Hyperparameters.} We first determine the two key hyperparameters of FaME: $f$, which controls the guidance strength, and $\tau$, which specifies when the FaME guidance begins during the denoising process. In practice, an extensive hyperparameter search is unnecessary, as our method demonstrates strong robustness and stability. The results are presented in Table~\ref{tab: hyperparameter}. We compute overall metrics rather than class-wise ones. Specifically, we sample 50{,}000 images (i.e., 50 images per class) to evaluate metrics such as FID. For the quality score $S_\text{Q}$, we calculate the score for each image individually and report the average as the final result.

The baseline model is DiT-XL/2 with a CFG scale of 1.5, whose performance is reported in the first row. In FaME, we adopt the same CFG scale. Under the setting $f = 0.02$, $\tau = 0.3$, the FID score even improves, and the quality score $S_\text{Q}$ increases significantly, so we adopt this configuration as our default. In particular, if diversity is not a concern or FID does not need to be considered, we can simply set $f = 0.1$ and $\tau = 1.0$ to further improve image quality.

\textbf{Quantitative Results.} To compare our method with the latest models quantitatively---including DiT-XL/2~\cite{dit}, SiT-XL/2~\cite{sit}, VAR-d30~\cite{var}, REPA (based on SiT-XL/2)~\cite{repa} and LightningDiT~\cite{vavae}---we evaluate these models using widely adopted metrics such as FID, IS, precision and recall. Additionally, we assess image quality using two representative IQA models: Q-Align~\cite{qalign} for computing $S_\text{Q}$ and HyperIQA~\cite{hyperiqa} for computing $S_\text{H}$. All models adopt CFG for generation. 

To evaluate our proposed FaME method, we apply it to both DiT-XL/2 and SiT-XL/2 and report the results in Table~\ref{tab: result}. Compared to the original DiT-XL/2 and SiT-XL/2, our method has minimal impact on metrics like FID, but leads to significant improvements in image quality scores. This demonstrates that FaME can effectively enhance baseline models and push their performance toward the best in terms of perceptual quality. Although the numerical gains in image quality scores may appear modest, we will shortly provide qualitative comparisons that reveal the substantial visual improvements achieved by our method.

\textbf{Qualitative Results.} To provide an intuitive understanding of the improvements brought by FaME, we visualize generation results across various classes. We use DiT-XL/2 with a CFG scale of 1.5 to generate images as the baseline. For a fair and stable comparison under the ODE sampling method, each pair of images, generated by CFG and by FaME, shares the same initial noise. We \emph{randomly} select outputs from each class and the results are shown in Figure~\ref{fig: visual-compare}. A clear improvement in visual quality can be observed. Images generated with DiT-XL/2 and CFG alone often exhibit noticeable distortion and blurriness, whereas applying FaME significantly enhances the clarity and realism of the generated samples.

In fact, some images generated by DiT-XL/2 with CFG are even visually unpleasant, an issue that is not uncommon in modern generative models. In particular, even advanced methods such as REPA and LightningDiT, which achieve FID scores lower than those of the validation set (i.e., real images), sometimes produce similarly undesirable images. This observation leads to a conjecture: current generative models can produce low-quality or distorted images that surprisingly do not worsen FID scores; in some cases, they may even improve them by increasing sample diversity or recall. Thus, metrics such as FID may not fully capture perceptual quality, highlighting the importance of qualitative and perceptual evaluation.

\textbf{Text-to-image Results}. Our method also has the potential to be applicable to text-to-image (T2I) generation. Traditional negative prompts require carefully crafted textual descriptions, which often fail when the prompt is ambiguous or unseen during training. In contrast, FaME uses negative images, which are easier to specify and more robust. We apply FaME to SDXL~\cite{sdxl}, and as shown in Figure~\ref{fig: t2i}, it effectively prevents the model from generating results similar to the negative samples. This demonstrates FaME’s potential beyond class-conditional generation.

\section{Conclusion and Discussion}

In this work, we were concerned on the community's over-reliance on distribution-based metrics such as FID and revisited the role of CFG in diffusion-based generation. Although CFG contributes to improved performance on distribution-based metrics such as FID and IS, these metrics somietimes do not align with human perception of image quality, which we believe may be more important.

Thus, we proposed FaME, a simple and efficient framework that leverages failure cases as negative guidance to steer the sampling process toward visually higher-quality regions. Without any retraining or additional inference cost, FaME provided consistent perceptual improvements across all classes and highlights a new direction for guidance design in generative models. On the same time, FaME achieved roughly the same FID score.

Despite its effectiveness, FaME has its limitations, for which we will focus on two aspects in our future work. First, FaME relies on the pre-selected set of representatives of failure modes, which may limit its generalization. Future work includes dynamically learning or adapting failure signals during sampling, extending FaME to more complex settings such as text-to-image generation, video generation, and exploring joint optimization of perceptual quality and diversity under user-defined preferences. Second, FaME relies on existing perceptual quality metric, while such metrics are far from being perfect. We also aim at finding a metric that aligns better with human perception across various visual generation domains and applications, to push the generative community toward being more human-centric.

\newpage
\clearpage
\bibliography{FaME}


\end{document}